\documentclass[review]{elsarticle}
\usepackage{lineno,hyperref}
\modulolinenumbers[1]
\journal{87th EAGE Annual Conference \& Exhibition, Aberdeen, UK, June 2026}
\date{}
\usepackage[utf8]{inputenc}
\usepackage{amsmath,amsthm,amsfonts,amssymb}
\usepackage{enumitem}
\usepackage{graphicx}
\usepackage{color}
\usepackage{float}
\usepackage{placeins}
\usepackage{multirow}
\usepackage{subcaption}
\usepackage{algorithm}
\usepackage{algpseudocode}
\usepackage{afterpage}
\usepackage{geometry}
\usepackage{float}

\graphicspath{ {./}}
\begin{document}
\makeatletter
\def\ps@pprintTitle{%
  \let\@oddhead\@empty
  \let\@evenhead\@empty
  \def\@oddfoot{\hfill\footnotesize\itshape
       \ifx\@journal\@empty\else\@journal\fi\hfill}%
  \let\@evenfoot\@oddfoot}
\makeatother

\begin{frontmatter}

\title{Property-Constrained 3D Porous Media Reconstruction from 2D Images via Conditional Generative Adversarial Networks}

\author[address1]{A. Sadeghkhani\corref{mycorrespondingauthor}}
\ead{alisdkhani@gmail.com}
\author[address1]{B. Bennett}
\author[address1]{A. Rabbani}
\cortext[mycorrespondingauthor]{Corresponding author}

\address[address1]{School of Computer Science, University of Leeds, Leeds, UK}

\begin{abstract}
This study presents a conditional Generative Adversarial Network (cGAN) framework for generating 3D porous media volumes with controlled porosity, trained exclusively on 2D thin section images. The key innovation lies in combining property-conditioned generation with 2D-to-3D reconstruction, eliminating the need for expensive 3D training data while maintaining control over petrophysical properties. The framework employs a hybrid architecture with a 3D generator and 2D discriminator, where multi-axis slice extraction enables learning 3D-consistent structures from 2D training data. Porosity labels are extracted using an Enhanced U-Net segmentation model. The methodology was demonstrated on two carbonate samples with different lithologies: dolomite-anhydrite and pure dolomite. Results show that the framework successfully generates realistic 3D volumes capturing lithological features such as anhydrite inclusions and fine crystalline textures. Porosity control achieved an $R^2$ of 0.93, with mean absolute errors of 0.019 and 0.010 for the heterogeneous and homogeneous samples, respectively.
\end{abstract}

\end{frontmatter}

\section{Introduction}

Three-dimensional characterisation of porous media microstructure is essential for understanding fluid flow behaviour in subsurface applications, including carbon storage, geothermal energy extraction, and groundwater management \citep{Blunt2013}. While advanced imaging technologies such as micro-CT and FIB-SEM can provide detailed 3D pore-scale data, these methods are constrained by high costs, limited sample availability, and practical acquisition challenges \citep{Cnudde2013}. In contrast, two-dimensional images remain widely available and cost-effective, yet they provide only partial information about the inherently three-dimensional pore network structure.

Recent advances in deep learning, particularly Generative Adversarial Networks (GANs), have demonstrated remarkable capabilities in generating realistic porous media images \citep{Goodfellow2014, Feng2020}. Several studies have explored 2D-to-3D reconstruction approaches, where 3D volumes are generated from 2D training data by exploiting statistical similarities between orthogonal cross-sections \citep{Feng2020, Shams2021}. However, most existing methods focus on unconditional generation without explicit control over petrophysical properties, limiting their practical utility for engineering applications that require specific property values.

Conditional GANs (cGANs) address this limitation by incorporating additional input parameters to guide the generation process \citep{Mirza2014}. Recent work has demonstrated property-controlled 3D porous media reconstruction, achieving precise control over porosity and other characteristics \citep{Zhou2023, Zheng2022}. However, these methods rely on 3D training data obtained from micro-CT imaging. Extending property-conditioned generation to 3D volumes while training exclusively on 2D images remains a significant challenge.

This study presents a slice-based cGAN framework for generating 3D porous media volumes (256\textsuperscript{3} voxels) with controlled porosity, trained using only 2D thin section images. The key innovation lies in combining property-conditioned generation with 2D-to-3D reconstruction, enabling control over petrophysical properties without requiring 3D training data. This is achieved through a hybrid architecture employing a 3D generator with a 2D discriminator, where multi-axis slice extraction enables the network to learn 3D-consistent structures from 2D training data. The methodology is demonstrated using carbonate thin section samples from a reservoir formation \citep{Rabbani2017}.

\section{Method and Theory}

\begin{figure}[H]
 \centering
 \includegraphics[width=0.7\textwidth]{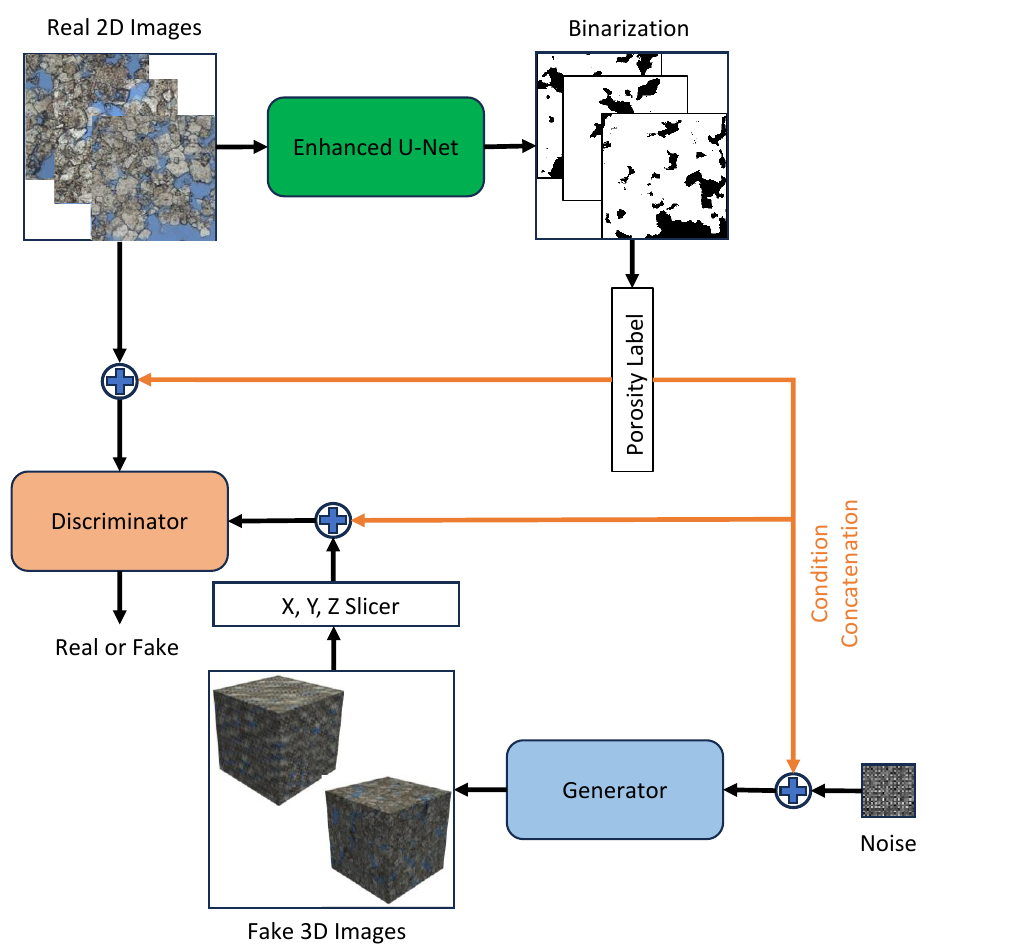}
 \caption{Overview of the proposed framework. Real 2D thin section images are binarised using an Enhanced U-Net to extract porosity labels. The generator receives noise concatenated with target porosity to produce 3D volumes, which are sliced along X, Y, and Z axes. The discriminator evaluates extracted slices against real 2D images, both conditioned on porosity labels.}
 \label{fig:framework}
\end{figure}

The proposed framework combines property-conditioned generation with 2D-to-3D reconstruction through an asymmetric architecture consisting of a 3D generator and a 2D discriminator (Figure~\ref{fig:framework}). The workflow comprises three main components including porosity label extraction, conditional 3D generation, and multi-axis discriminator training.

Porosity labels are extracted from training images using an Enhanced U-Net segmentation model \citep{Sadeghkhani2025}, which binarises the images by distinguishing pore spaces (identified by blue-dyed epoxy resin) from solid matrix with high accuracy (Dice coefficient = 0.95). The computed porosity values serve as conditioning labels for the GAN training.

The generator $G$ receives a latent noise vector concatenated with a target porosity value $c$, and produces a $256^3$ voxel RGB volume. The discriminator $D$ operates on 2D slices and evaluates whether they exhibit realistic pore structures consistent with the specified porosity. During training, multiple 2D slices are extracted from generated volumes along three orthogonal axes ($X$, $Y$, $Z$) and compared against real 2D training images, both paired with their corresponding porosity labels. The training follows the standard conditional adversarial objective:

\begin{equation}
L = \mathbb{E}_{x}[\log D(x|c)] + \mathbb{E}_{z}[\log(1 - D(G(z|c)|c))]
\end{equation}
where $x$ represents real 2D images, $z$ is the latent vector, and $c$ is the porosity condition.

The model was trained on thin section images from two core samples obtained from a carbonate reservoir formation \citep{Rabbani2017}. Both samples exhibit crystalline fabric with intercrystalline pore types; the first comprises dolomite-anhydrite lithology with distinctive anhydrite mineral inclusions, while the second consists of pure dolomite with fine-grained crystalline texture.

\section{Results}

\begin{figure}[!htb]
 \centering
 \includegraphics[width=0.85\textwidth]{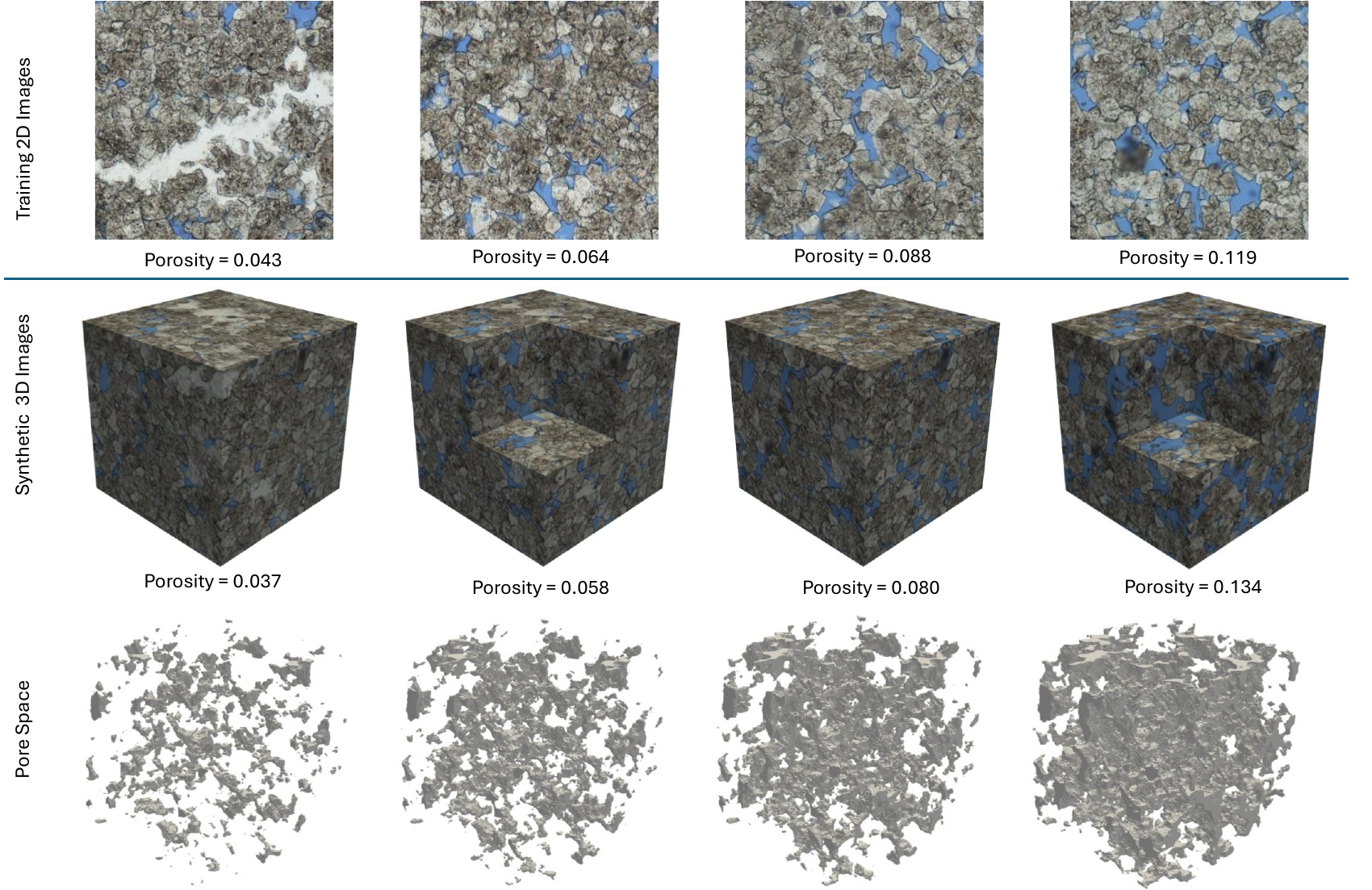}
 \caption{Results for Sample 1 (dolomite-anhydrite). Top row: training 2D thin section images with corresponding porosity values. Middle row: generated 3D volumes at specified target porosities. Bottom row: extracted pore space from generated volumes showing increasing pore volume with porosity.}
 \label{fig:results_sample1}
\end{figure}

The framework was evaluated on two carbonate samples with distinct lithological characteristics. Figure~\ref{fig:results_sample1} presents results for the dolomite-anhydrite sample, where the generated 3D volumes successfully reproduce the characteristic light-coloured anhydrite mineral inclusions observed in the training images. These impermeable anhydrite patches are consistently represented throughout the generated volumes, demonstrating that the network captures lithological heterogeneity beyond simple pore-solid distributions.

\begin{figure}[!htb]
 \centering
 \includegraphics[width=0.85\textwidth]{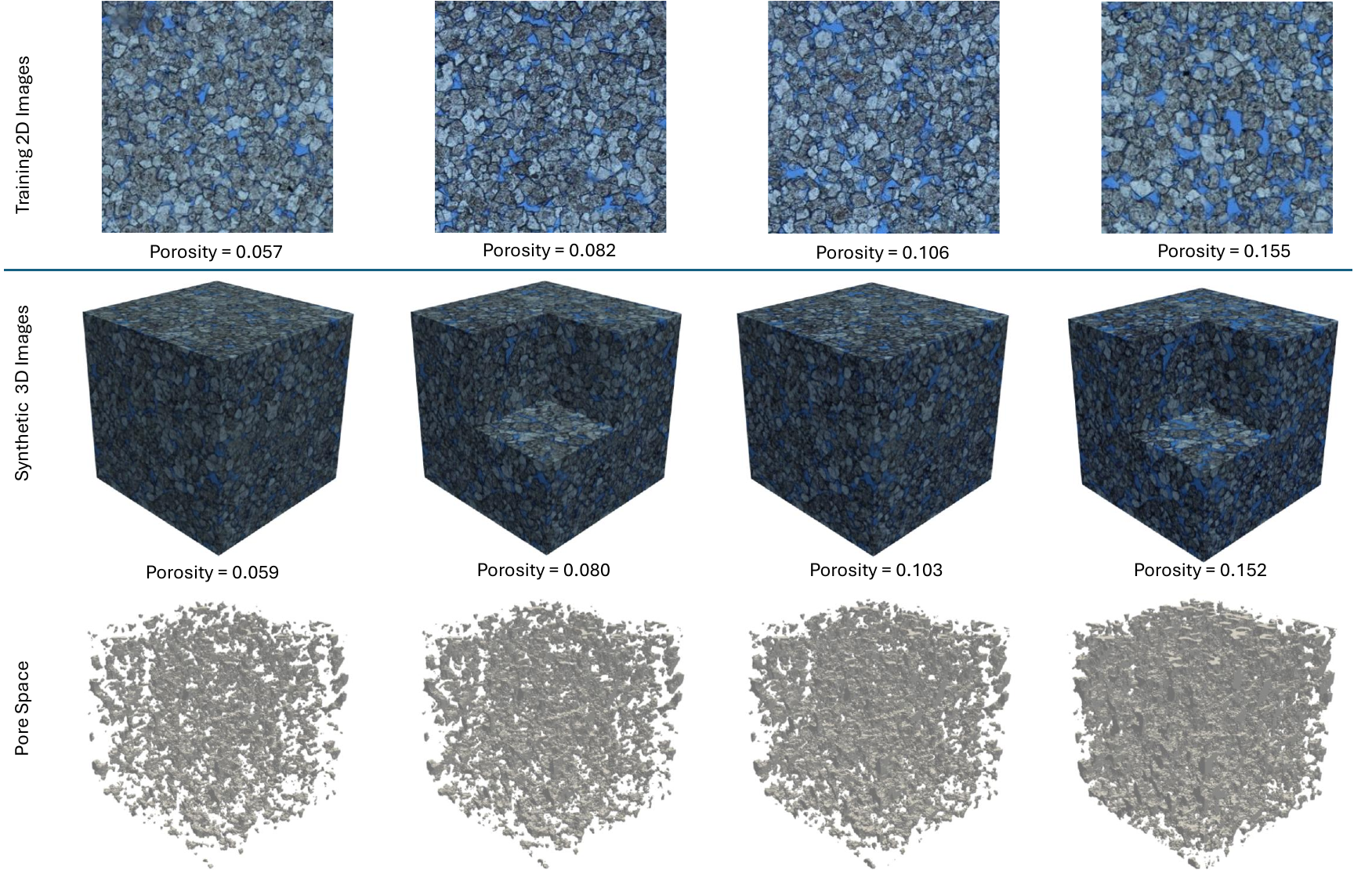}
 \caption{Results for Sample 2 (pure dolomite). Top row: training 2D thin section images with corresponding porosity values. Middle row: generated 3D volumes at specified target porosities. Bottom row: extracted pore space from generated volumes.}
 \label{fig:results_sample2}
\end{figure}

Figure~\ref{fig:results_sample2} shows results for the pure dolomite sample, which exhibits a finer crystalline texture with smaller, more uniformly distributed pores. The generated 3D volumes faithfully reproduce this fine-grained pattern, with the pore space visualisations confirming the characteristic small-scale porosity distribution observed in the training images.

Porosity control accuracy was assessed by comparing target values with measured porosities from generated volumes. The framework achieved an overall coefficient of determination ($R^2$) of 0.93, indicating strong correlation between target and generated porosity values. Sample 1 (dolomite-anhydrite) achieved a mean absolute error (MAE) of 0.019, while Sample 2 (pure dolomite) demonstrated higher precision with an MAE of 0.010. The improved accuracy for Sample 2 is attributed to its more homogeneous texture compared to the heterogeneous dolomite-anhydrite lithology. Both samples show that generated porosity increases progressively with target values, confirming effective property conditioning.

\section{Conclusions}
This study presented a conditional GAN framework for generating 3D porous media volumes with controlled porosity, trained exclusively on 2D thin section images. The key contribution lies in combining property-conditioned generation with 2D-to-3D reconstruction, eliminating the need for 3D training data while maintaining control over petrophysical properties.

Results demonstrated successful generation of realistic 3D volumes for two carbonate samples with different lithologies. The framework captures not only porosity variations but also lithological features such as anhydrite inclusions and fine crystalline textures. Porosity control achieved an $R^2$ of 0.93, with MAE values of 0.019 and 0.010 for the heterogeneous and homogeneous samples respectively.

Future work will include quantitative validation through pore size distribution analysis, two-point correlation functions, and flow simulation studies, as well as extending the framework to additional petrophysical properties.

\end{document}